\documentclass[letterpaper, 10 pt, conference]{ieeeconf}  

\IEEEoverridecommandlockouts                              

\usepackage{url}
\usepackage{graphicx}
\usepackage{booktabs}
\usepackage{romannum}
\usepackage{amsmath}
\usepackage{amsfonts}
\usepackage{multirow}
\usepackage[table,xcdraw]{xcolor}
\usepackage{array}
\usepackage{cleveref}
\usepackage{float}
\usepackage[export]{adjustbox} 
\usepackage{algorithm}
\usepackage{balance}
\usepackage{cite}
\usepackage{algpseudocode} 
\usepackage{longtable}
\usepackage{booktabs}
\usepackage[german]{babel}
\usepackage[utf8]{inputenc}
\newcommand{\squeezeup}{\vspace{-2.5mm}}
\UseRawInputEncoding

\title{\LARGE \bf
Combining Learning from Demonstration with Learning by Exploration \\to Facilitate Contact-Rich Tasks 
}

\author{Yunlei Shi$^{1,2}$, Zhaopeng Chen$^{2,1}$, Yansong Wu$^{3}$, Dimitri Henkel$^{2}$,\\ Sebastian Riedel$^{2}$, Hongxu Liu$^{3}$, Qian Feng$^{3,2}$, Jianwei Zhang$^{1}$
\thanks{*This research has received funding from the German Research Foundation (DFG) and the National Science Foundation of China (NSFC) in project Crossmodal Learning, DFG TRR-169/NSFC 61621136008, partially supported by European projects H2020 STEP2DYNA (691154) and ULTRACEPT (778602).}
\thanks{{$^{1}$TAMS (Technical Aspects of Multimodal Systems), Department of
Informatics, Universit\"at Hamburg}, {$^{2}$Agile Robots AG}, {$^{3}$Technische Universit\"at M\"unchen.}}}

\begin{document}
\maketitle
\thispagestyle{empty}
\pagestyle{empty}


\begin{abstract}
Collaborative robots are expected to work alongside humans and directly replace human workers in some cases, thus effectively responding to rapid changes in assembly lines. Current methods for programming contact-rich tasks, particularly in heavily constrained spaces, tend to be fairly inefficient. Therefore, faster and more intuitive approaches are urgently required for robot teaching. This study focuses on combining visual servoing-based learning from demonstration (LfD) and force-based learning by exploration (LbE) to enable the fast and intuitive programming of contact-rich tasks with minimal user efforts. Two learning approaches were developed and integrated into a framework, one relying on human-to-robot motion mapping (visual servoing approach) and the other relying on force-based reinforcement learning. The developed framework implements the noncontact demonstration teaching method based on the visual servoing approach and optimizes the demonstrated robot target positions according to the detected contact state. The developed framework is compared with two most commonly used baseline techniques, i.e., teach pendant-based teaching and hand-guiding teaching. Furthermore, the efficiency and reliability of the framework are validated via comparison experiments involving the teaching and execution of contact-rich tasks. The proposed framework shows the best performance in terms of the teaching time, execution success rate, risk of damage, and ease of use. 
\end{abstract}

\section{INTRODUCTION}
The ability to rapidly setup and reprogram newly-introduced products in factories is an increasingly essential requirement for adaptive robotic assembly systems \cite{kramberger2016learning}, \cite{sloth2020towards}. Position-controlled robots can handle known objects within well-structured assembly lines with high efficiency and achieve highly accurate position control. However, they require considerable setup time and tedious reprogramming to fulfill new tasks, and cannot adapt to any unexpected variations in assembly processes \cite{zhu2018robot}.
\begin{figure}[htbp]
\centerline{\includegraphics[width=8.6cm,height=4.9cm]{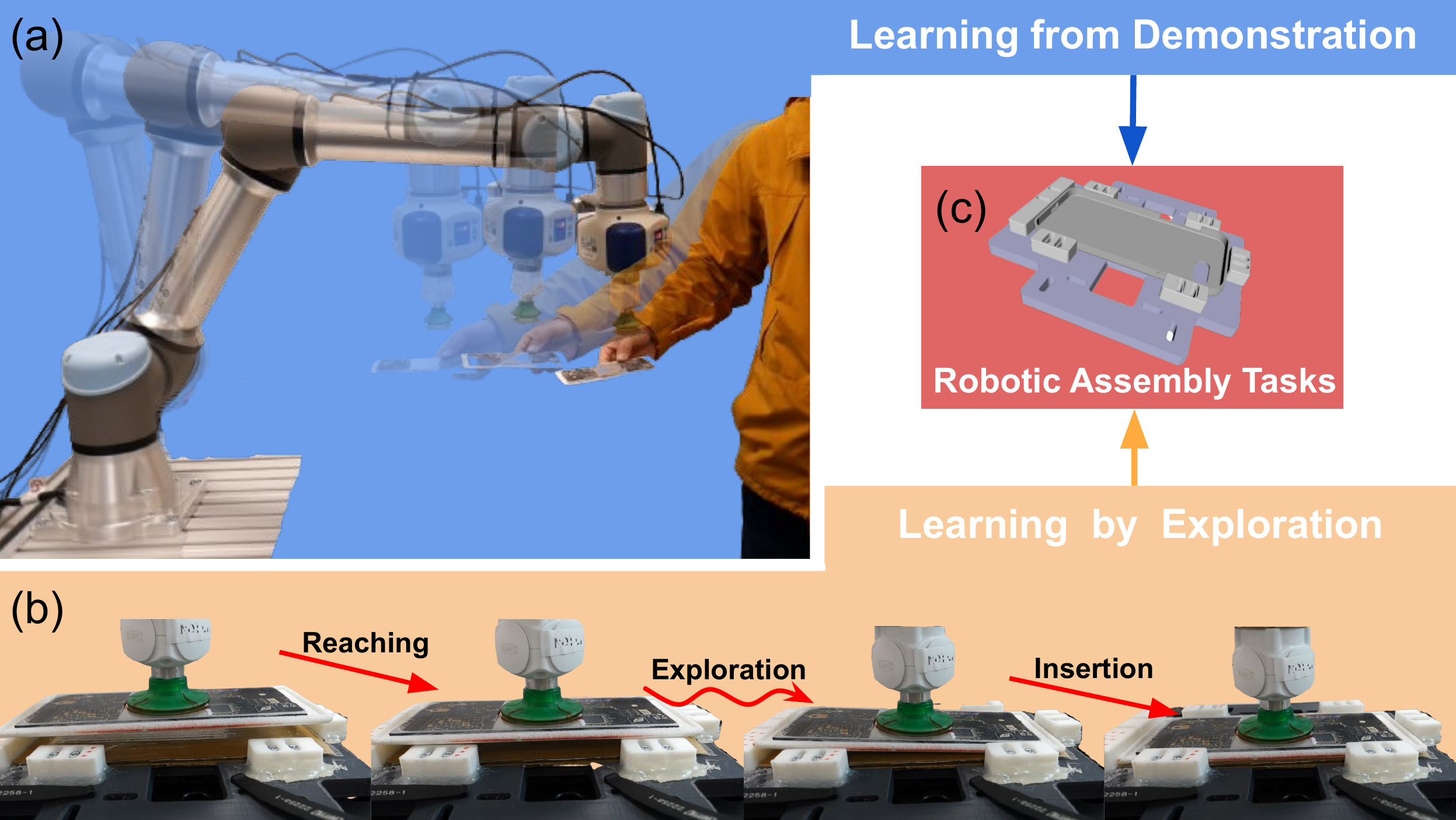}}
\caption{A robot arm and a suction gripper performing a contact-rich tending task. (a) Gross motion is learned from human demonstration. (b) Fine motion is learned from exploration. (c) Example of a contact-rich tending task}
\label{fig:system framework}
\squeezeup
\squeezeup
\end{figure}


Collaborative robots offer the promise of closing the gap between onerous reprogramming and unexpected variations by combining the capabilities of position-controlled robots with dexterity and flexibility. For example, the hand-guiding method enables unskilled users to interact with collaborative robots and facilitates quick programming \cite{safeea2017end}. However, during assembly line reconfiguration, a long time is still required to remove and reinstall the robot arms and various attachments.
Mobile manipulators (where robotic arms are mounted on mobile bases) were introduced to expand the productivity and adaptive capacity of manufacturing automation, particularly during the setting up phase when production lines must be reconfigured \cite{marvel2013towards}. Because mobile manipulators can only be placed beside production lines and cannot be installed on production lines as collaborative robots, which occupying space previously provided for human workers. However, programming a robot in a constrained space is very difficult \cite{lee2020robot}. Overall, the ease-of-programming has been identified as an open challenge in robot assembly \cite{sloth2020towards}, \cite{marvel2018multi}.

Additionally, collaborative robots equipped with force control functions can perform certain hybrid position/force operations for contact-rich tasks \cite{albu2007unified}, \cite{inoue2017deep}, \cite{gaz2019dynamic}, \cite{lee2018making}; however, their effectiveness and variation adaptive capacity in assembly processes are still unsatisfactory \cite{luo2019reinforcement}, \cite{schoettler2019deep}. 
Herein, we propose an intuitive programming method to decrease the setup time of mobile manipulators and introduce a reinforcement learning (RL) algorithm to overcome unexpected variations in assembly tasks (\Cref{fig:system framework}). Our primary contributions are: 

\begin{itemize}
\item \textbf{C1:} An approach is presented that learns the trajectories of robots from demonstrations based on visual servoing for fast, easy, and accurate robot setup in heavily constrained spaces. 
\item \textbf{C2:} A region-limited residual RL (RRRL) policy based on force-torque information is trained to overcome pose uncertainty.
\item \textbf{C3:} \textbf{C1} and \textbf{C2} are combined into a robot tending skill and compared with two most commonly used baseline techniques using an UR5e robot.
\end{itemize}

The rest of the paper is organized as follows. In \Cref{RELATED WORK AND BACKGROUND}, we provide an overview of the development of industrial mobile manipulators and robot programming methods, as well as their pros and cons. Moreover, the force controller is briefly introduced. In \Cref{PROBLEM STATEMENT AND METHOD OVERVIEW}, we outline some key problems of assembly tasks and corresponding ideas of our method. Details on our proposed method are provided in \Cref{POLICY DESIGN}, and a quantitative experiment of our methods is presented in \Cref{EXPERIMENTS}. Experimental results are presented and discussed in \Cref{RESULTS AND DISCUSSION}. Finally, potential future research areas are suggested in \Cref{CONCLUSIONS}.

\section{RELATED WORK AND BACKGROUND}\label{RELATED WORK AND BACKGROUND}

\subsection{Industrial Mobile Manipulator}\label{IndustrialMobileManipulator}
Mobile manipulator systems, also known as mobile manipulator hybrid robots\footnote{http://std.samr.gov.cn/} or collaborative cells\footnote{https://blog.robotiq.com/}, were designed to reduce the setup time of industrial robots; all the required task devices are integrated into a mobile platform, and it is therefore a ``plug and play'' solution.
Various mobile manipulators have been developed, such as Rob@Work \cite{helms2002rob}, Little Helper\cite{hvilshoj2009mobile}, KMR iiwa \cite{wurll2018production}, KUKA flexFELLOW\footnote{www.kuka.com} and MiR UR\footnote{https://mobile-industrial-robots.com/}, which are the most widely-deployed industrial mobile manipulators. However, the traditional magnetic tracking approach utilized by commercial mobile manipulators (i.e., automatic guided vehicle) can be accompanied by position uncertainty problem. This does not satisfy the requirements of many industrial applications, particularly when mobile manipulators are required to move between picking and placing locations.

\subsection{LfD Under Visual Guidance}\label{programming and assembly by visual guidance}
LfD has recently been recommended as an effective technique for accelerating the programming of learning processes, spanning from low-level control to high-level assembly planning \cite{kruger2014technologies}, \cite{davchev2020residual}. Guiding robots by means of visual feedback \cite{lee2019making}, \cite{zheng2017peg}, \cite{huang2017vision} during assembly tasks is an effective way to overcome position uncertainties introduced by mobile manipulators. For robotic assembly tasks, performing high precision measurements is important. However, visual errors can be introduced by lenses and the imaging sensors, as well as the calibration of intrinsic and extrinsic parameters \cite{li2019survey}. Some researchers believe that humans should focus more on execution tasks than vision sensors \cite{li2019survey}. Based on this notion, other approaches have been developed, such as intelligent assembly algorithms, in an effort to lower the necessity of vision sensors for given tasks.
\subsection{LbE for Contact-rich Tasks}\label{Reinforcement Learning for Contact-rich Tasks}

LbE has been suggested as an effective method for reducing the programming time, and recent studies have introduced artificial intelligence methods into robotics \cite{kramberger2016learning}, \cite{zhu2018robot}, \cite{al2019robotic}, \cite{bogue2014role}. 
\begin{figure}[htbp]
\centerline{\includegraphics[width=8.5cm,height=4.0cm]{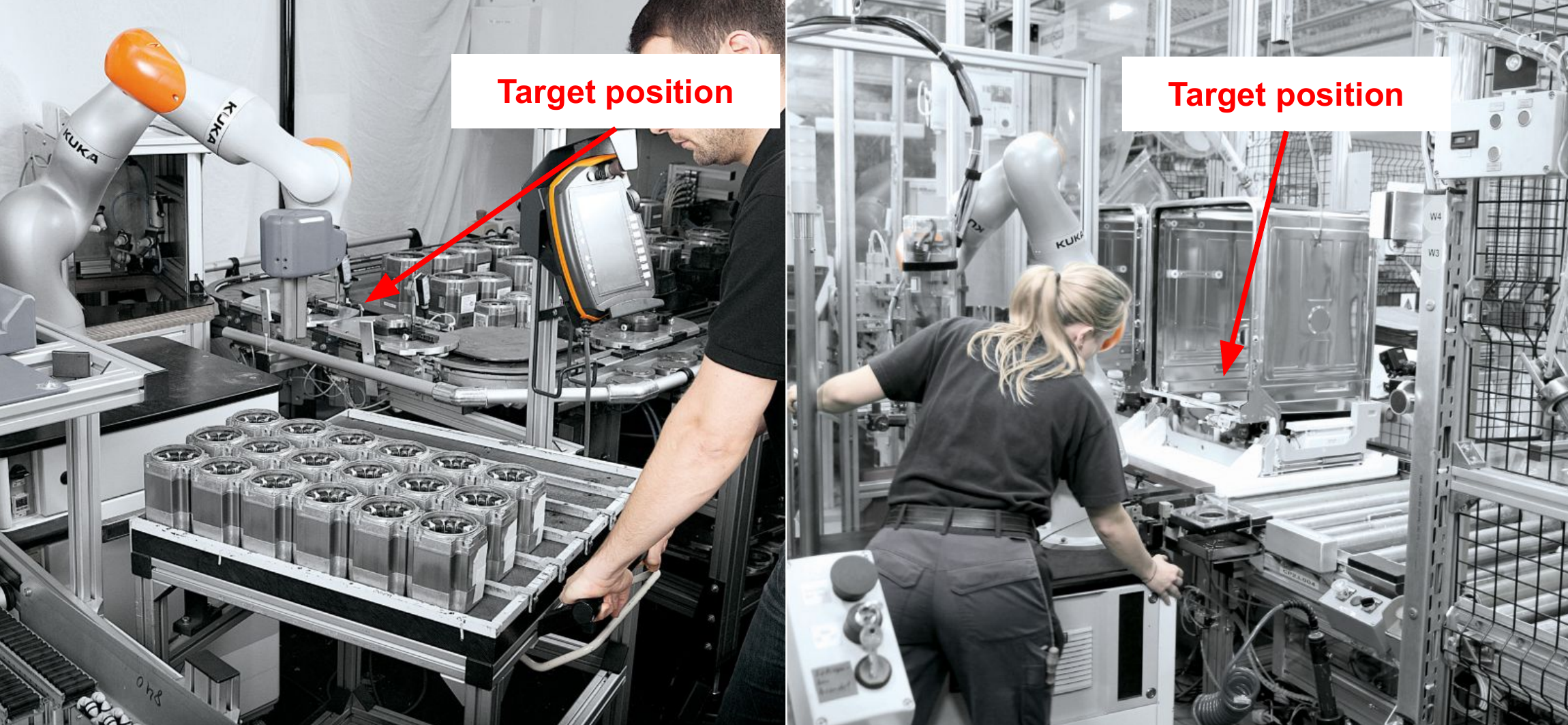}}
\caption{Examples of heavily constrained spaces in real factories $^\mathrm{5}$.} 
\label{fig:Heavily constrained space}
\squeezeup
\squeezeup
\end{figure}
\footnotetext[5]{www.kuka.com/en-de/products/robot-systems/industrial-robots/lbr-iiwa} Moreover, RL offers a set of tools for designing sophisticated robotic behaviors that are difficult to engineer. RL and its derivative methods have previously been successfully used to address various robotic manipulation problems \cite{lee2018making}, \cite{luo2018deep}, \cite{inoue2017deep}, \cite{luo2019reinforcement}, \cite{lee2020guided}, \cite{feng2020center}. 
Exploration behavior entails interactions between robots and their operational environment. Therefore, a force or impedance controller is required. 
\subsection{Operational Space Force Control}

Interaction control has attracted widespread scholarly interest following pioneering works on impedance/admittance control and passivity by Hogan \cite{hogan1985impedance} and Colgate \cite{colgate1988robust}. However, the first mention of admittance control concepts dates back to the literature by Whitney \cite{whitney1977force}, where these concepts were used to address hard contact issues for industrial manipulation functions and indirect force control purposes. Overall, operational space force control has proven the effectiveness of RL-based assembly tasks \cite{inoue2017deep}, \cite{luo2019reinforcement}.

\section{PROBLEM STATEMENT AND METHOD OVERVIEW}\label{PROBLEM STATEMENT AND METHOD OVERVIEW}

\subsection{Programming with Mobile Manipulators}\label{Fast and easy Programming with mobile manipulator}
\subsubsection{Problem details}
Mobile manipulators can considerably reduce robots and devices installation time \cite{matheson2019human}. Programming based on demonstration approaches has been proposed to address variations in geometry and configurations for assembly, placement, handling, and picking tasks \cite{matheson2019human}, which can reduce the programming time and user training requirements \cite{zhu2018robot}. The use of mobile manipulators introduces a \textbf{positioning error at the $\pm$5 mm level} \cite{su2018positioning}, and errors as small as $\pm$1 mm can induce large \textbf{huge contact forces and consistent failures} in typical assembly tasks \cite{schoettler2019deep}. In the present study, we address the more typical cases of mobile bases that involve the repositioning of mobile manipulators according to task requirements.
\begin{figure*}[htbp]
\centerline{\includegraphics[width=18cm,height=6.0cm]{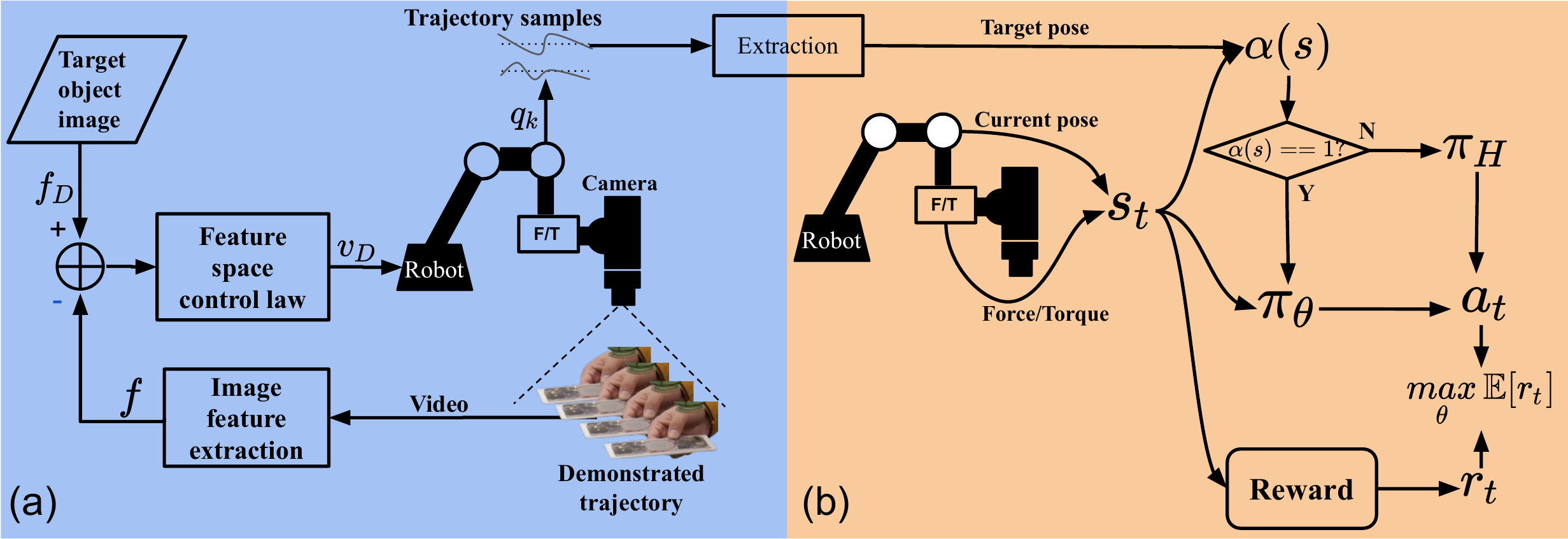}}
\caption{Combination of LfD policy (a) and RRRL policy (b)}
\label{fig:CombinationPolicies}
\squeezeup
\squeezeup
\end{figure*}

Teach pendants are still used for precision positioning (position and orientation of the end effector (EE)) in many tasks \cite{safeea2017end}. However, these devices limit the intuitiveness of teaching processes and are time-consuming. Hand-guiding is a typical physical contact kinesthetic teaching solution, where programming is embodied using demonstration concepts, enabling users to quickly and intuitively program robots. However, it has drawbacks in terms of accuracy, locational separations, and operations involving dangerous objects \cite{zhu2018robot}.
Moreover, neither the hand-guiding nor the teach pendant programming methods can compensate for the positioning errors that accompany mobile units \cite{su2018positioning}, and can result in the generation of huge contact force that can damage objects.  

Additionally, the base of a mobile manipulator requires considerable space within work cells. \Cref{fig:Heavily constrained space} shows some real-world factory examples of constrained spaces. Sometimes, a user must teach a robot with a highly \textbf{awkward body posture} owing to anthropometric limitations \cite{lee2020robot}. Moreover, the delicate movement required by a user may be difficult to realize due to the resistance of robots in the drag mode \cite{lee2020robot}. Hence, because of these two issues, the use of hand-guiding for accurate assembly is rather difficult and yields low quality (e.g., excessive contact force and low accuracy). 

\subsubsection{Method overview}
We propose a method that is simple and fast to implement and reduce the physical contact force to solve the problems described in \Cref{Fast and easy Programming with mobile manipulator}. 
We introduce visual guiding into the teaching phase as shown in \Cref{fig:CombinationPolicies}(a). The proposed visual guiding teaching is performed using two steps:
\begin{itemize}

\item \textbf{Grasping pose definition:} First, a user guides the robot to achive a grasping pose under the target object frame. Then, the robot moves up and uses an eye-in-hand camera to capture a photo of the object as a reference.
\item \textbf{Trajectory generation:} Second, the robot follows the moving object (the object can be moved by the user) to achieve a new target pose using a vision-based control algorithm (e.g., visual servoing \cite{kragic2002survey},\cite{hutchinson1996tutorial}), and records the entire moving trajectory. 
\end{itemize}



Comparing to utilizing a global camera \cite{wei2018design}, the method proposed in this paper uses an eye-in-hand camera to effectively avoid the occlusion of the target, and ensure that the robot follows the object to achieve the proper trajectory, thus achieves better performance in real industrial scenarios. Trajectory errors have little effect on final assemblies because uncertainties can be generally ignored in gross motion planning, and a fine motion planner can solve uncertainties during the assembly process \cite{gottschlich1994assembly}.

\subsection{Assembly Under Position Uncertainty }
\subsubsection{Problem details}
Visual sensors can be used for target recognition, pose estimation, measurement, and positioning using traditional methods \cite{li2019survey}. However, for visual sensors, lenses, imaging sensors and the calibration of intrinsic/extrinsic parameters considerably influence the precision of visual guidance, as well as reflection, shadow, and occlusion may fail to extract the edges and features of objects owing to lights changes and object textures \cite{li2019survey}.

\subsubsection{Method overview}
Residual RL \cite{schoettler2019deep}, \cite{johannink2019residual} is a novel method that exploits the efficiency of conventional controllers and the flexibility of RL. Residual RL attempts to introduce prior information in an RL algorithm to accelerate the training process, rather than performing random explorations from scratch. For example, the estimated position can be set as prior information and even can have errors. 
The guided uncertainty-aware policy optimization (GUAPO) \cite{lee2020guided} showed better performance than residual RL, soft actor critic (SAC), and pure model-based method such as deep object pose estimator (DOPE). However, the force and torque information is not considered in the GUAPO policy. This information can provide observations regarding current contact conditions between objects and their environments for accurate localization \cite{lee2019making}. Moreover, it can ensure manipulation safety \cite{luo2019reinforcement}. However, pure force-based learning policies may lead to substantial deviations from goals and reduce learning efficiency. Thus, we combine the ``region limitation'' idea from GUAPO and the residual RL policy to develop a force-based approach called ``\textbf{R}egion-limited \textbf{R}esidual \textbf{RL}'' (\textbf{RRRL}) (\Cref{fig:CombinationPolicies}(b)). 
In the RRRL policy, the rough target pose is obtained using by the teaching phase as the residual part and a function $\alpha(s) = 1[s \in \mathbb{S}_u]$ is used to switch between the fixed $\pi_H(s)$ and the parametric $\pi_{\theta}(s)$ policies\cite{lee2020guided}:
\begin{equation}
\setlength{\abovedisplayskip}{3pt}
\setlength{\belowdisplayskip}{3pt}
  \pi(a|s) =(1-\alpha(s)) \cdot \pi_H(a|s) + \alpha(s) \cdot \pi_{\theta}(a|s).
  \label{equation:RRRL}
\end{equation}
$\mathbb{S}_u$ is the region containing the goal position with uncertainty.
Because force control is more safe and reliable in the fine motion/manipulation phase than position control and impedance control in assembly tasks \cite{inoue2017deep}, the RRRL policy $\pi_{\theta}(s)$ takes the operational force controller as the desired force/torque in operational spaces. thus, our goal is to maximize expected return (i.e., $\underset \theta\max \mathbb{E}[r_t]$) through the RRRL policy. The fixed policy $u_H$ is used to move the object back to the initial target pose when the function $\alpha(s) = 0$.

\section{POLICY DESIGN}\label{POLICY DESIGN}

The processes in LfD can be divided into three steps: observing, representing, and reproducing an operation \cite{zhu2018robot}.
We perform our entire policy following these processes, and then add the RRRL policy at the end of the operations. Our method comprises two learning policies: 

\begin{itemize}

\item \textbf{Lfd policy}: The robot learns gross motions via human demonstrations, wherein a human demonstrates the target object images and grasping position to the robot.
\item \textbf{RRRL policy}: The robot learns fine motions based on the RRRL policy, as described in \Cref{algorithm:RRRL}. The RRRL policy can be trained in advance to save the setup time.

\end{itemize}

First, we define the terminology and notation required to represent the coordinate transformations. We represent the task space of the robot as $\mathcal{T}$, which constitutes a set of positions and orientations that can be attained by the robot EE (i.e., suction gripper). $\mathcal{T}$ is a smooth $m$-manifold in which $m$ = 6 and $\mathcal{T=}$ $SE^3 = \mathcal{R}^3 \times SO^3$.
The superscripts/subscripts of the coordinate frames are listed in \Cref{tab:coordinate frames}.

\begin{table}[H]
    \normalsize
    \centering
	\caption{coordinate frames}
	\label{tab:coordinate frames}
	\begin{tabular}{|c|l|}
		\hline
		$e$ & Coordinate frame attached to the robot EE \\ \hline
		$c$ & Coordinate frame of the camera  \\\hline
		$b$ & Base coordinate frame of the robot \\\hline
		$o$ & Coordinate frame attached to the target object \\ \hline
	\end{tabular}
\end{table}

\subsection{LfD Policy} 
In our policy, we equip an eye-in-hand camera to avoid the occlusions caused by the robot's links and other industrial devices \cite{zhu2018robot}, \cite{lippiello2007position} during the demonstration. The relative homogeneous transformation $^{e}\boldsymbol{x}_c$ and intrinsic camera parameters are determined by means of the hand/eye calibration method \cite{tsai1989new}.

In the demonstration phase, teaching was categorized into the three following steps:

\noindent 1) Robot EE was moved to the \textbf{desired grasping pose (DGP)}, which was then recorded as $^{b}\boldsymbol{x}_{DGP}$.

\noindent 2) Robot EE was moved to the \textbf{desired visual servoing pose (DVSP)}, which was recorded as $^{b}\boldsymbol{x}_{DVSP}$; here, the object must be kept in view of the camera. The \textbf{first reference photo (RF1)} was captured, and a fixed \textbf{relative pose (RP)} was calculated as follows: 
\begin{equation}\label{relativepose}
\setlength{\abovedisplayskip}{3pt}
\setlength{\belowdisplayskip}{3pt}
\begin{split}
&^{c}\boldsymbol{x}_{o} =(^{c}\boldsymbol{x}_{e}) (^{DVSP}\boldsymbol{x}_{DGP}) =  (^{c}\boldsymbol{x}_{e}) (^{b}\boldsymbol{x}_{DVSP})^{-1}  (^{b}\boldsymbol{x}_{DGP}),\\
\end{split}
\end{equation}
where $^{c}\boldsymbol{x}_{o}$ is the coordinate transformation of the object frame $o$ with respect to the camera frame $c$.

\begin{algorithm}
    \caption{\textbf{RRRL}}
    \algorithmicrequire{\;Model based policy $\pi_H$, learning frequency $C_1$, and target action-value update frequency $C_2$. }
    \begin{algorithmic}[1]
        \State Initialize replay memory $\mathcal{H}$ to capacity $N$
        \State Initialize action-value function $Q$ with random weights $\theta$
        \State Initialize target action-value function $Q_{target}$ with weights $\theta^- = \theta$
        \For {episode = 1 to $M$}
        \State Sample state $s_0$
            \While {NOT EpisodeEnd}
                \State Calculate $\alpha(s)$ using \Cref{alpha calculation}
                \State Select action $a_H$ from $\pi_H(s_t)$
                \State With probability $\epsilon$, select a random action $a_{RL}$
                \State Otherwise select $a_{RL} \sim \pi_\theta(s_t)$
                \State Obtain action $a_t =(1-\alpha)*a_H + \alpha * a_{RL}$
                \State Execute $a_t$, and observe reward $r_t$ and state $s_{t+1}$
                \State Store transition $(s_t, a_t, r_t, s_{t+1})$ in $\mathcal{H}$ with priority $p_t = max_{i<t} p_i$
                \For {$j$ = 1 to $C_1$}
                    \State Sample minibatch of transitions with priority from $\mathcal{H}$
                    \State Update transition priority
                    \State Update $\theta$ using the method proposed in \cite{schaul2015prioritized}
                \EndFor
                \State In each $C_2$ step, reset $Q_{target} = Q$
            \EndWhile
        \EndFor
    \end{algorithmic}
\label{algorithm:RRRL}
\end{algorithm}
\noindent 3) The visual servoing strategy \cite{hutchinson1996tutorial} was activated, and the following system constraints \cite{chan2011constrained} were applied to the robot during the teaching process:

\begin{align} \label{systemConstraints}
    \mathbf q & \in\mathbb{Q}_c,\\
    \label{equation3}\mathbf q & \in{[\mathbf {q}^{min}, \mathbf {q}^{max}]},\qquad \qquad \mathbf q^{min}, \mathbf q^{max} \in \mathbb R^N, \\
    \label{equation4}\mathbf {\dot{q}} & \in{[\mathbf {\dot{q}}^{min}, \mathbf {\dot{q}}^{max}]},\qquad \qquad \mathbf  {\dot{q}}^{min}, \mathbf {\dot{q}}^{max} \in \mathbb R^N, \\
    \mathbf {q} & {_{k+1}} = \mathbf {{q}}_k +\delta \mathbf {{\dot{q}}}_k,
\end{align} 

where $\mathbb{Q}_c$ is the set of configurations that do not cause any part of the arm to collide with obstacles that are difficult to model. \Cref{equation3} and \Cref{equation4} describe the robot's joint positions and velocity constraints, respectively.
The object was then moved from the \textbf{DGP} to the \textbf{desired final pose (DFP)} by the user, the robot EE followed the trajectory $\mathbf {q}_k$ from the \textbf{DVSP} to the \textbf{DFP} under the aforementioned constraints. Further, the trajectory could be recorded. At \textbf{DFP}, the camera automatically captured the \textbf{second reference photo (RF2)}, and the \textbf{DFP} could be easily calculate at the end of the trajectory by \Cref{DFP}:
\begin{equation}\label{DFP}
\setlength{\abovedisplayskip}{3pt}
\setlength{\belowdisplayskip}{3pt}
\begin{split}
&DFP = (^{b}\boldsymbol{x}_{e}) (^{e}\boldsymbol{x}_{c}) (^{c}\boldsymbol{x}_{o})\\
\end{split}
\end{equation}

In this study, the image-based visual servoing method was introduced based on specified observed feature positions (\Cref{fig:CombinationPolicies}(a)). With the human expert (user) in the teaching loop, the trajectory-based representation output using LfD could avoid collisions even if the robot operated in a heavily constrained operation space. Additionally, our noncontact guiding method can avoid the resistance force of the robot in the drag mode, making it easier and more accurate for the users to perform teaching. 

\subsection{RRRL Policy}  

In this section, we explain the RRRL algorithm in detail as shown in \Cref{fig:CombinationPolicies}(b) and \Cref{algorithm:RRRL} to learn the assembly tasks. 

\subsubsection{Limited Region and Policy Structure}
In most assembly tasks, the aim is to minimize the distance between objects and their goal positions \cite{luo2019reinforcement}. The limited region is used to constrain the exploration area. Many methods can describe the uncertainty region with respect to a nonparametric distribution \cite{lee2020guided} or a parametric equation. In this study, the Euclidean distance was used to indicate the switch signal between the fixed $\pi_H(s)$ and parametric $\pi_{\theta}(s)$ policies in the hybrid RRRL policy:
\begin{equation}\label{alpha calculation}
    \alpha(s)= 
\begin{cases}
    1, & \text{if}\left\| CP - DFP \right \|_2 < D \\
    0, & \text{otherwise,}\\
\end{cases}
\end{equation}

CP and DFP denote the current and desired final poses, respectively. D is an engineering hyperparameter determined based on experience, and we suggest that D should be at least twice the final positioning error introduced by the fixed policy $\pi_H(s)$. When $ \alpha(s)=1$, the \textbf{force-based} learning policy $\pi_{\theta}(s)$ is used, otherwise the \textbf{position-based} fixed policy $\pi_H(s)$ is used to the object back to its initial uncertain target pose. The same switch algorithm was used during the training and execution phases. A double DQN with proportional prioritization \cite{schaul2015prioritized} was selected as the learning policy $\pi_{\theta}(s)$ in this study.
\subsubsection{Action Design}
The actions in the assembly task can be either a position \cite{lee2019making}, \cite{lee2020guided} or force/torque command \cite{inoue2017deep}, \cite{luo2019reinforcement}. Because we aim to reduce the contact force between the object and environment to ensure safety, the force/torque command action in the operational space (i.e. under frame $x_e$) was selected \cite{luo2019reinforcement}.

Carefully exploiting the natural constraints in the design of the learning policy was essential for the assembly tasks considered herein. It is obvious that the task is simplified if the motion is constrained in ``wrong'' directions. We utilized comparative experiments to investigate the ways in which different force-based actions utilize natural constraints. We set the same initial positional error of $\delta{P} \in [2,4]$ mm in a random direction, and then performed a random strategy to select the actions. We tested each action 200 times with a maximum steps of 20 and a maximum force amplitude of 10N, the success rates of which are as follows:
\begin{table}[H]
\centering
\begin{tabular}{@{}|l|c|c|@{}}
\toprule
  & \textbf{Force control action}           & \textbf{Success rate} \\ \midrule
1 & Operational space controller \cite{luo2019reinforcement}             & 32\%                  \\ \midrule
2 & Fz with the force of another one dimension \cite{inoue2017deep} & 60\%                  \\ \midrule
3 & Fz with the forces of other two dimensions & 69\%                  \\ \bottomrule
\end{tabular}
\end{table}
\begin{figure}[htbp]
\centerline{\includegraphics[width=8.8cm,height=4cm]{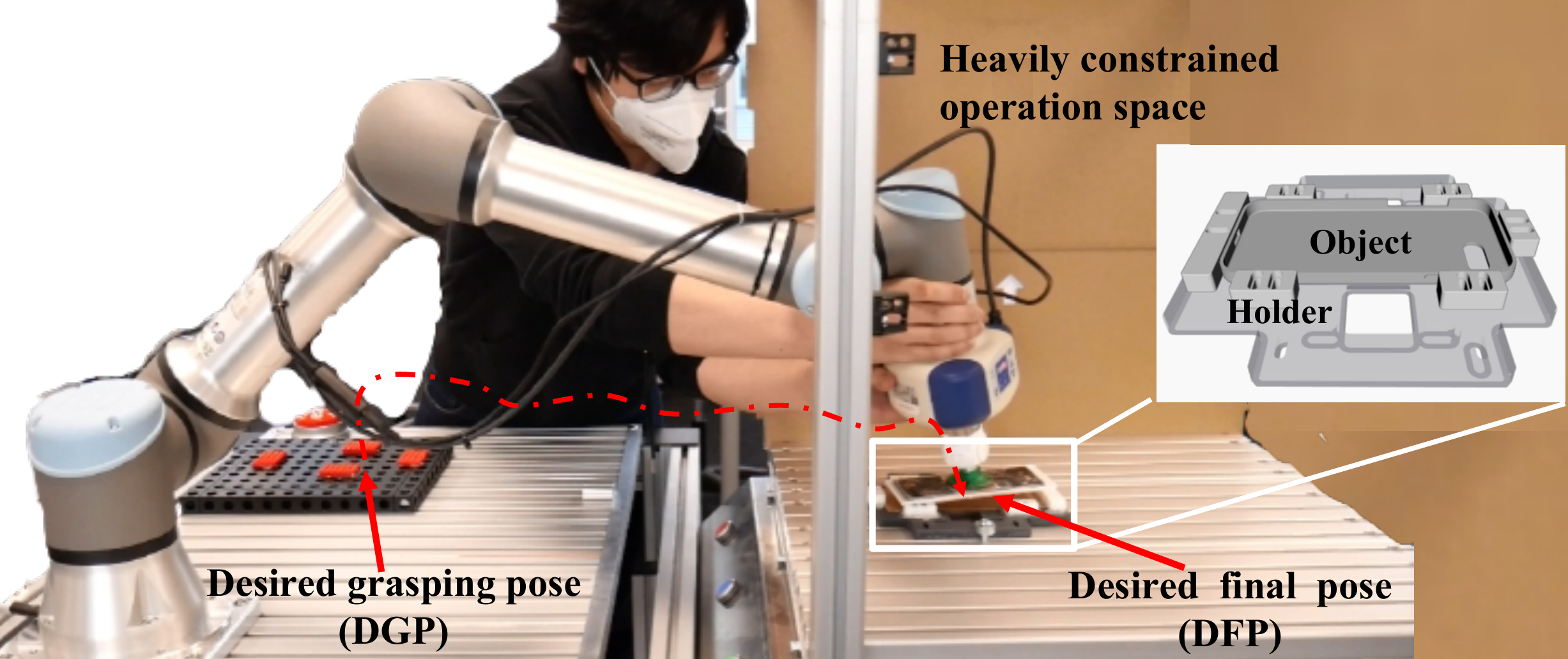}}
\caption{Hand-guiding teaching in a heavily constrained operation space. Both the user and robot arms are close to their operating boundaries (near singular configurations).}
\label{fig:setup}
\squeezeup
\squeezeup
\end{figure}
The third discrete action as \Cref{actions} performs the best in the comparison experiment:
\begin{equation}\label{actions}
\setlength{\abovedisplayskip}{3pt}
\setlength{\belowdisplayskip}{3pt}
\begin{split}
1:&[+^eF_x, +^eF_y, +^eF_z, 0, 0, 0]\\
2:&[+^eF_x, -^eF_y, +^eF_z, 0, 0, 0]\\
3:&[-^eF_x, +^eF_y, +^eF_z, 0, 0, 0]\\
4:&[-^eF_x, -^eF_y, +^eF_z, 0, 0, 0]\\
\end{split}
\end{equation}
Here, we set the orientation space as the position mode to utilize the flexibility of the suction cup. Then, all the torque commands in the operation space were set to 0 Nm, and all the force amplitudes were set to 10 N, which is half of force amplitudes set in\cite{inoue2017deep}.

\subsubsection{State Design}
Forces and torques feature the most direct information that characterizes contact states during an operation.Thus, the 6-dimensional force-torque vector $s = [F_x, F_y, F_z, M_x, M_y, M_z]$ under the robot EE's frame $x_e$ were sent to the RRRL network as the input state. The state sampling frequency was 10 Hz, which could guarantee the observation of the contact states.

\subsubsection{Reward Design}
We employed the precise target position of the hole as a reference for the reward during the learning phase. Unlike the execution phase, the precise desired position was easy to obtain because the robot exhibited high positional repeatability.
\begin{equation}
    r= 
\begin{cases}
    1-k_{steps}/k_{max},              & \text{success}\\
    -  \Vert CP - DFP \Vert_2 ,& \text{otherwise}.
\end{cases}
\end{equation}

\subsubsection{Parameters Setup}
In order to gain more contact experience, a positional error $\delta{P} \in [2,4]$ mm was added in a random direction during the training phase. The transitions $(s_t, a_t, r_t, s_{t+1})$ sampled from the environment were stored in a replay buffer \cite{schaul2015prioritized}. The size of the experience's replay memory $P_{replay}$ was 20, 000, the maximum number of training episodes $M$ was 200, and the maximum number of steps $k_{max} $ for the search phase was 50. The batch size $P_{batch}$ was set to 64 to select random experiences from $P_{replay}$, and the discount factor $\lambda $ was 0.5.

\section{EXPERIMENTS}\label{EXPERIMENTS}
\begin{figure}[htbp]
\centerline{\includegraphics[width=8.8cm,height=4.9cm]{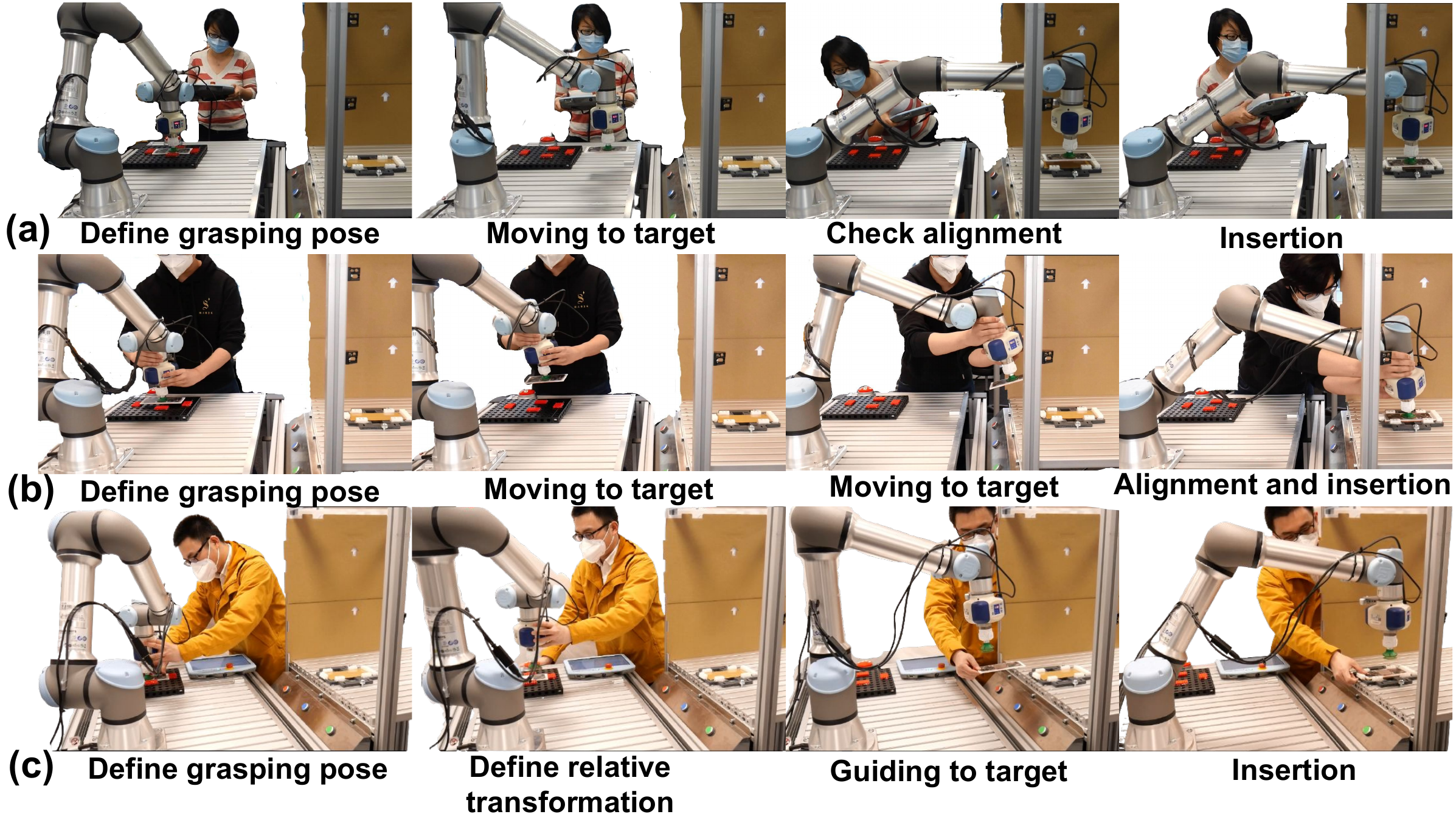}}
\caption{(a) Teach-pendant teaching: aligning the object with the target holder by eye is difficult. (b) Hand-guiding teaching: moving the robot EE with singular configurations and align with target holder are difficult. (c) Our LfD method teaching: contact-free guiding that is not physically demanding, and is easy to align without robot resistance.}
\label{fig:teachPhase}
\squeezeup
\squeezeup
\end{figure}
We evaluated our methods in the \textbf{teaching} and \textbf{execution} phases. In our experiments in both phases, we aimed to answer the following questions: (\textbf{Q1}) Can our method maintain fast and easy programming abilities even in constrained operation spaces? (\textbf{Q2}) Can our method retain the execution success rate against positional uncertainty? (\textbf{Q3}) Can our method reduce the risk of damage during operation of an object?

A UR5e robot \footnote{https://www.universal-robots.com/products/ur5-robot/} was used to implement our novel approach to facilitate a tending task. The UR5e features a 6-axis and 5 kg payload, a working radius of 850 mm. It is equipped with a 6-DOF force/torque sensor on the EE. UR5e robot uses admittance controller \cite{hogan1985impedance} to achieve operational space force control. A Schmalz CobotPump ECBPi suction cup was installed beyond the force/torque sensor in order to ensure the detection of the contact force with the environment. An Intel RealSense Depth Camera D435i was attached to the EE to conduct the visual servoing process. Our policy was run using a Dell Precision 5510 laptop, and the updated position was sent to the UR5e controller. We used ur-rtde \footnote{https://pypi.org/project/ur-rtde/} as the Python interface for controlling and receiving data from the UR robot. A 6-DoF ATI Axia80 force sensor \footnote{https://www.ati-ia.com/Products/ft/sensors.aspx} was mounted under the holder in order to measure the operating force, and a low-pass filter with a cutoff frequency of 9.37 Hz was used to mitigate the force noise.

\subsection{Baseline Techniques}
The commonly available methods for collaborative robots were selected as baselines \cite{lee2020robot}. We compared our proposed method with the following baselines:

\noindent \textbf{Baseline 1: Teach-pendant + spiral searching.} The UR5e teach pendant with a UR PolyScope GUI was held in one hand by a user, who pressed the on-screen buttons to map the rate control of the EE's translation and rotation in the task space $\mathcal{T}$ with the other. At the DFP, a spiral search function similar to that outlined in the literature \cite{park2020compliant} was added. 

\noindent \textbf{Baseline 2: Hand-guiding + spiral searching.} The UR5e ``Freedrive'' mode was utilized, and users physically grabbed and exerted force to move the robot arm using one or two hands. At the DFP, a spiral search function was also added.
\begin{figure}[htbp]
\centerline{\includegraphics[width=8.8cm,height=6.5cm]{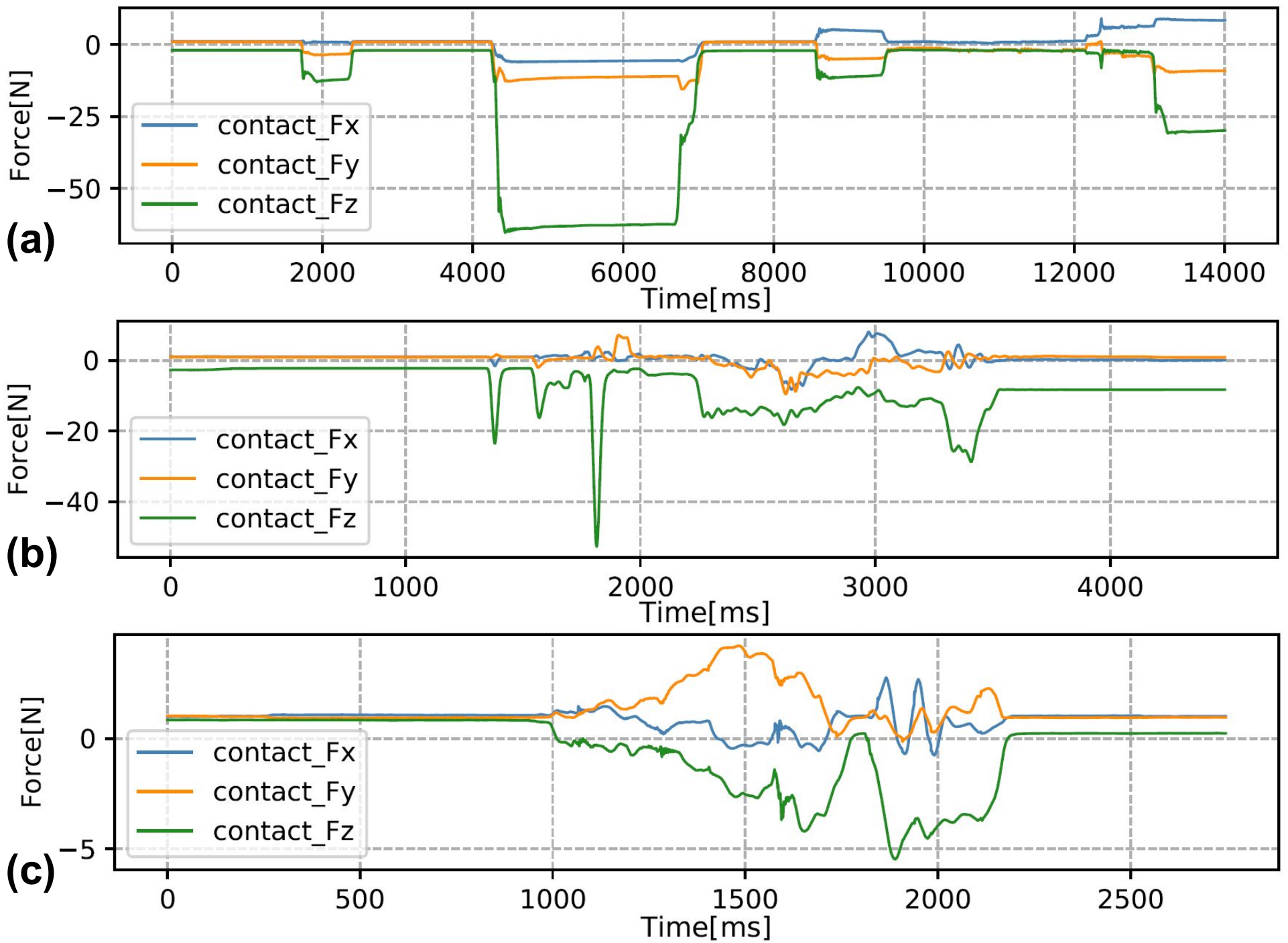}}
\caption{Contact force during teaching. (a) This curve shows the contact characteristics of the teach-pendant teaching method: the user uses an ``observe-move'' strategy and only makes adjustment after finding unsuccessful insertions, hence, the contact force is maintained during the observation. (b) The hand-guiding method: the object frequently collides with the holder when it is not inserted (impact force in the figure), and there is a continuous contact force after inserting the object. (c) Our method produces a small contact force (less than 5 N) in the teaching phase.}
\label{fig:teachForce}
\squeezeup
\end{figure}
\begin{figure}[htbp]
\centerline{\includegraphics[width=8.8cm,height=4.3cm]{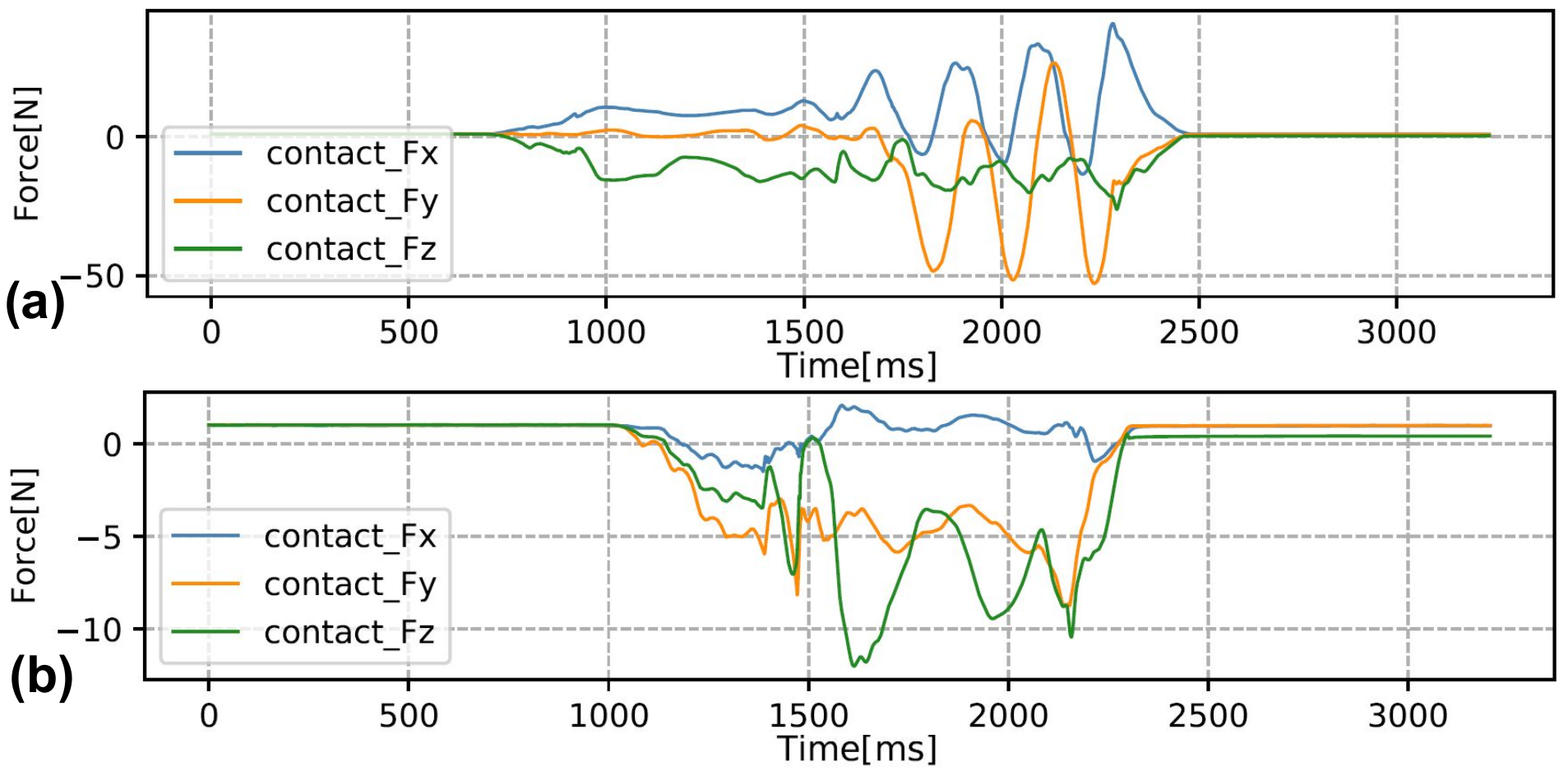}}
\caption{Contact force during execution. (a) Spiral exploration method: the first half of the curve (700-1500ms) is not yet constrained by the holder, and the contact force is small and stable. After 1600ms, the object is inserted into the holder, immediately generating a larger contact force. Due to the accuracy issue of the UR5e force sensor, the spiral movement generates a larger contact force than the stopping threshold. (b) RRRL method: the maximum contact force is around 10N because of the force action amplitude limitation.}
\label{fig:executeForce}
\squeezeup
\squeezeup
\end{figure}
\subsection{Task Setup}
A tight clearance machine tending task (similar to the assembly task) was used to evaluate our method and the two baselines (\Cref{fig:setup}) . One holder was installed in an opaque box to simulate a situation in which the field of view in the production line is obscured and an object is inserted into the holder. The users were provided a brief tutorial and allowed to practice until they felt ready. A group of four able-bodied volunteers (1 female, 3 male, aged: 24 to 35) participated in the experiment.

\subsection{Measurement and Evaluation}
\begin{itemize}
\item \textbf{Teaching time:} The task's completion time was measured as the time costed by the user to move the robot from the DGP to the DFP (\Cref{evaluation in the teaching phase}).
\item \textbf{Execution success rate:} After the teaching phase was completed, the trajectory of the demonstration was executed and the insertion success rate was tested. Contrary to the \textbf{``perfect''} group, an error of $\delta{P} \in [2,4]$ mm in a random direction was added on DFP to simulate the pose uncertainties in the \textbf{``uncertainty''} group (\Cref{evaluation in the execution phase}). 
\item \textbf{Risk of damage:} The risk of damage caused by the operating force was ignored in previous studies \cite{lee2018making}, \cite{luo2018deep}, \cite{luo2019reinforcement}, \cite{lee2020guided}, \cite{park2020compliant}. In this study, the maximum absolute contact force during contact operation was employed to evaluate the risk of damage (\Cref{evaluation in the execution phase}).
\end{itemize}

\section{RESULTS AND DISCUSSIONS}\label{RESULTS AND DISCUSSION}

\begin{table}[]
\caption{evaluation in the teaching phase}
\label{evaluation in the teaching phase}
\centering
\begin{tabular}{|c|c|c|c|}
\toprule
\textbf{Teaching phase} &Time cost  & Maximum contact force \\ \midrule
Teach-pendant           &60--120 s                 &15--50 N           \\ \midrule
Hand-guiding            &\textbf{15--42 s}                  &30--60 N           \\ \midrule
Our method              &\textbf{23--30 s}                  &\textbf{3--10 N}           \\ \bottomrule
\end{tabular}
\end{table}

\begin{table}[]
\caption{evaluation in the execution phase}
\label{evaluation in the execution phase}
\centering
\begin{tabular}{@{}|c|c|c|c|@{}}
\toprule
\multirow{2}{*}{\textbf{Execution phase}}                                   & \multicolumn{2}{c|}{\textbf{Success rate}} & \multirow{2}{*}{\textbf{\begin{tabular}[c]{@{}c@{}}Maximum \\ contact force\end{tabular}}} \\ \cmidrule(lr){2-3}
                                                                            & \textbf{Perfect}   & \textbf{Uncertainty}  &                                                                                            \\ \midrule
Only teach-pendant                                                          & 55/100             & 17/100                & 15 N                                                                                        \\ \midrule
Only hand-guiding                                                           & 33/100             & 5/100                 & 15 N                                                                                        \\ \midrule
\begin{tabular}[c]{@{}c@{}}Teach-pendant + \\ spiral searching\end{tabular} & 69/100             & 47/100                & 35 N                                                                                        \\ \midrule
\begin{tabular}[c]{@{}c@{}}Hand-guiding + \\ spiral searching\end{tabular}  & 51/100             & 33/100                & 35 N                                                                                        \\ \midrule
Our method                                                                  & \textbf{95/100}             & \textbf{91/100}                & \textbf{15 N}                                                                                        \\ \bottomrule
\end{tabular}
\end{table}

Overall, the results of 12 group robot teaching and 1000 group robot execution were obtained.

The teaching phase test scenarios can be observed in \Cref{fig:teachPhase}, and the results are presented in as \Cref{evaluation in the teaching phase}. Although it is more generalized, our method features a similar time cost to the hand-guiding method. Using the hand-guiding method, male volunteers always required less teaching time than females owing to physical demands. All the volunteers noted that the robot ``required excessive force to move, particularly near the boundaries''. Our method does not require physical contact with the robot, therefore, it is not physically demanding. All the volunteers required considerably longer teaching times when the teach-pendant teaching method was used. considerable amount of time was spent to align the object with the target, and the tricky sight angle made this even more difficult. In contrast, our method does not require a human to align the object with the target, but enables the robot arm to actively track the object to attain the target position. Thus, the setup is performed quickly. Hence, the answer to \textbf{Q1} is yes. Furthermore, our method produces minimal contact force on the environment (\Cref{fig:teachForce}), as our schematic approach is identical to the human tending one. The other two methods resulted in considerably less transparency during the interactions with the environment owing to the resistance of the robot arm itself or the inability to interact with the environment in terms of force.

The execution phase results are presented in \Cref{evaluation in the execution phase}. The success rate of our method far exceeded those of the other reference baselines. We found that \textbf{the elasticity of the suction cup} had a major influence on the accuracy of the demonstrated target position. The contact force at the end of the demonstration could induce the \textbf{deformation of the suction cup} and thus affect the actual target position. Using our method, the robot EE did not contact the target environment at the end of the teaching phase; hence, there was no contact force and therefore no deformation of the suction cup. Finally, high target position accuracy was achieved. Additionally, the RRRL policy could determine force actions based on the contact state of the object and holder, thus greatly improving the insertion success rate. Our method can guarantee small contact forces owing to the amplitude limitation of the force actions (\Cref{fig:executeForce}(b)). Therefore, the answers to \textbf{Q2} and \textbf{Q3} are also yes.
\section{CONCLUSIONS}\label{CONCLUSIONS}
Collaborative robots are rapidly making advances in new work environments. However, there is a control and teaching bottleneck in the programming of contact-rich tasks.

In this study, we combined visual servoing based LfD and force-based LbE to facilitate the rapid and intuitive execution of the assembly tasks that require minimal user expertise, involvement, and physical exertion. The efficiency of the proposed method was validated via a series of experiments that involved the execution of a tending task using a robot arm and a suction cup system.

In a challenging setting designed to simulate heavily constrained operation spaces, which is very common in actual factories, experiments that compared our method with two commonly used baselines, namely teach pendant-based and hand-guiding teaching, were performed. Our method realized the best feedback in terms of both subjective and objective evaluations

We noted the negative effect of the elasticity of the suction cups on the accuracy of the position demonstrations. In the future, we hope to investigate ways to utilize this elasticity using a learning approach. We also plan to analyze more contact-rich tending tasks and more types of grippers to refine our method and improve its generalizability.

\section*{Acknowledgment}
We would like to thank Chengjie Yuan, Chunyang Chang, Jingjing Sun, and Alexander Daimer for their contributions as volunteers and for their valuable feedback during the discussion of the experimental protocol. We also thank mechanical engineer Ningxin Lu for building the hardware mockup system.

\clearpage
\balance

\bibliographystyle{IEEEtran}
\bibliography{references}

\end{document}